\documentclass[letterpaper, 10 pt, conference]{ieeeconf}  

\usepackage[pdftex]{graphicx}
\usepackage{caption}
\usepackage{subcaption}

\usepackage[T1]{fontenc}
\usepackage{CJKutf8}
\usepackage[english]{babel}

\usepackage{tabularx,booktabs}
\usepackage{multirow}
\usepackage{hhline}

\usepackage{lipsum}

\usepackage{amsmath, mathtools}

\usepackage{tikz}
\usepackage{textcomp}

\newcommand\copyrighttext{%
  \footnotesize \textcopyright 2020 IEEE. To appear in IV, 2020. Personal use of this material is permitted.  Permission from IEEE must be obtained for all other uses, in any current or future media, including reprinting/republishing this material for advertising or promotional purposes, creating new collective works, for resale or redistribution to servers or lists, or reuse of any copyrighted component of this work in other works.}
\newcommand\copyrightnotice{%
\begin{tikzpicture}[remember picture, overlay]
\node[above, align=left, outer sep=0pt, yshift=15pt] at (current page.south) {\fbox{\parbox{\dimexpr\textwidth-\fboxsep-\fboxrule\relax}{\copyrighttext}}};
\end{tikzpicture}%
}

\newcommand\urltext{%
  \footnotesize \url{https://sites.google.com/view/iv20-mha-prediction/}}
\newcommand\displayurl{%
\begin{tikzpicture}[remember picture, overlay]
\node[above, anchor=north, yshift=-126pt] at (current page.north) {\urltext};
\end{tikzpicture}%
}

\IEEEoverridecommandlockouts                              

\makeatletter
\let\NAT@parse\undefined
\makeatother
\usepackage{hyperref}  

\title{\LARGE \bf
Multi-Head Attention based Probabilistic Vehicle Trajectory Prediction
}

\author{Hayoung Kim$^{1}$, Dongchan Kim$^{1}$, Gihoon Kim$^{1}$, Jeongmin Cho$^{1}$ and Kunsoo Huh$^{1}$
\thanks{$^{1}$Hayoung Kim, Dongchan Kim, Gihoon Kim, Jeongmin Cho and Kunsoo Huh are with the Department of Automotive Engineering, Hanyang University, Seoul, 04763, Republic of Korea; authors email {\tt\small \{hayoung.kim, jookker, rlgns4861, jo87964, khuh2\}@hanyang.ac.kr} }%
}

\begin{document}

\begin{CJK}{UTF8}{mj}

\maketitle 

\thispagestyle{empty}
\pagestyle{empty}

\copyrightnotice
\vspace{-\baselineskip}

\displayurl
\vspace{-\baselineskip}

\begin{abstract}
This paper presents online-capable deep learning model for probabilistic vehicle trajectory prediction. We propose a simple encoder-decoder architecture based on multi-head attention. The proposed model generates the distribution of the predicted trajectories for multiple vehicles in parallel. Our approach to model the interactions can learn to attend to a few influential vehicles in an unsupervised manner, which can improve the interpretability of the network. The experiments using naturalistic trajectories at highway show the clear improvement in terms of positional error on both longitudinal and lateral direction. 

\end{abstract}

\section{Introduction}

One of the most difficult problems in autonomous driving is to perceive and understand their surroundings. For safe and efficient decision making, it is necessary to accurately forecast the future trajectories of surrounding vehicles. However, it is challenging to accurately predict the trajectory. This is because the inherent uncertainty exists in the future trajectory itself and the behaviors of the surrounding vehicles affect to each other. To tackle these challenges, the prediction model should consider both interaction among vehicles and their uncertainty. 

In this paper, we propose a probabilistic model for vehicle trajectory prediction, which can consider the interaction among surrounding vehicles and the road environment. In order to model the vehicle interaction, multi-head attention architecture in Transformer \cite{vaswani2017attention} is utilized and it is considered as a major breakthrough in field of natural language processing. When humans drive, they internally predict future trajectories of the surrounding vehicles. Instead of predicting trajectories for all the surrounding vehicles, humans focus on a small number of influential vehicles to plan safe and efficient trajectory. The proposed prediction model is motivated by this characteristic of the human driver. We want to make the model learn to attend a few influential vehicles naturally. In addition, by encoding lane features using the attention mechanism, the prediction model reflects the contextual information of the road environments. It helps the model to better predict the future trajectories of surrounding vehicles. 

To evaluate the proposed model, naturalistic trajectories recorded at highways are used. In the experiments, the proposed method is compared with the existing methods, where the model jointly learn the distribution of the future trajectories and the interactions. 

The proposed model have several attractive properties for vehicle trajectory prediction as follows:

\begin{itemize}
    \item \textit{Interpretability}: The use of multi-head attention improves the interpretability of the neural network because the model can learn the social relations of neighboring vehicles in an unsupervised manner.
    
    \item \textit{Scalability}: As the output dimension of multi-head attention is flexible to the number of the vehicles, the proposed network can be extended to very dense traffic scenarios. The network is tested in an autonomous vehicle platform with surrounding vehicles less than 30. The average computation time is 50ms.
    
    \item \textit{Accuracy}: The proposed method is verified by using the naturalistic trajectory data in highway, and the results show the better performance than the existing methods in terms of positional error.

\end{itemize}

\begin{figure*}
    \centering
    \includegraphics[width=0.97\linewidth]{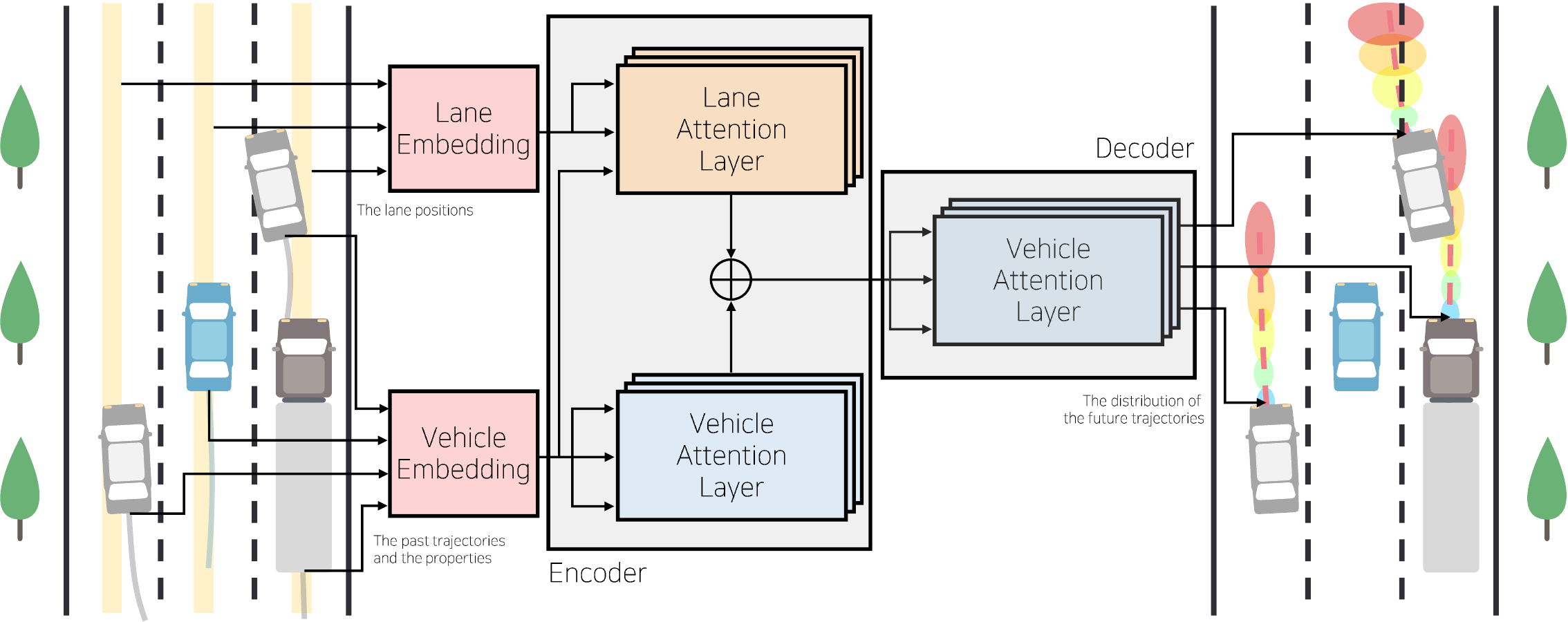}
    \caption{Proposed prediction architecture. The road on the left denotes the input of the prediction model, which consists of the past trajectories of surrounding vehicles, $\mathbf{X}$, and the lane information, $\mathbf{I}$. The road on the right denotes the output of the prediction model, which is the distribution of the future trajectories, $P(\mathbf{Y}|\mathbf{X}, \mathbf{I})$.}
    \label{fig:architecture}
\end{figure*}

\section{Related Work}

Classically, there were several researches for trajectory prediction assuming that the vehicle moves according to a certain motion model (e.g. CTRA; Constant Turn Rate and Acceleration) \cite{berthelot2011handling, polychronopoulos2007sensor}. In \cite{xie2017vehicle}, the authors integrated the motion model and the maneuver based model using Interactive Multiple Model (IMM) filters, which improved the accuracy of longer term prediction. In \cite{wiest2012probabilistic}, the future trajectory was predicted based on motion pattern extracted from the past trajectory, which used Gaussian Mixture Model (GMM) to consider uncertainty.

Recently, deep learning techniques have succeeded in the field of natural language processing, which is closely related to real life, and their techniques have been widely used to design the models for trajectory prediction \cite{altche2017lstm}, \cite{park2018sequence}. In \cite{altche2017lstm}, lateral position and longitudinal velocity were predicted using Long-Short Term Memory (LSTM) \cite{hochreiter1997long}. The LSTM used the current state values such as position, velocity, distance from the preceding vehicle, and time-to-collision (TTC). Authors in \cite{park2018sequence} used the encoder-decoder structure of Sequence-to-Sequence \cite{sutskever2014sequence} for trajectory prediction. After the past trajectory was encoded using LSTM, future trajectory sequence was generated using LSTM decoder. The main contribution of the prediction framework is that beam search can generate multiple future trajectories with high probability. However, since the future trajectory is predicted by the occupancy grid representation, it inherently contains an error corresponding to the size of the grid.

All the vehicles on the road maintain a certain social distance to avoid collisions with each other. For this reason, the importance of predicting the future trajectory by reflecting the interaction among the vehicles is more emphasized rather than independently predicting the trajectory of each vehicle \cite{lee2017desire, feng2019vehicle, li2019grip}. A framework for generating diverse trajectory samples with Conditional Variational Auto-Encoder (CVAE) \cite{NIPS2015_5775} and refining the trajectory using the inverse optimal control was introduced in \cite{lee2017desire}. In the refinement process, the interaction of surrounding vehicles is considered using social pooling \cite{alahi2016social}. In \cite{feng2019vehicle}, the authors proposed a prediction model which used the behavior level intention as a condition. Graph Convolutional Network (GCN) \cite{kipf2016semi}, which is an emerging topic in deep neural network, has been applied with grid representation in the trajectory prediction in order to model the interactions of close vehicles in \cite{li2019grip}. These studies used the semantic information of the vehicles (e.g. front vehicle, front left vehicle, rear right vehicle etc.) for interaction modeling, or the maximum distance was manually set where the interaction occurs. However, the suggested method improves the prediction performance by learning the attention distribution via multi-head attention in an unsupervised manner, which makes the model to focus on vehicles with intimate social interaction.

\section{Proposed Prediction Model}

It is very difficult to consider all the interactions with surrounding vehicles in the autonomous driving. As the number of surrounding vehicles increases, the complexity of the interactions increases more than exponentially. Interestingly, human drivers can reduce complexity by focusing on vehicles with intimate social interaction even if there are many surrounding vehicles. This motivates us to apply multi-head attention, which is used in Transformer \cite{vaswani2017attention}, for the vehicle trajectory prediction problem. Although studies on the relations among surrounding vehicles was conducted, the relations had to set manually as a semantic positional information (\cite{xie2017vehicle, scheel2019attention, dong2019interactive}), rather than learning them automatically. In this section, the proposed model structure is described consisting of encoder and decoder with two attention layers.

\subsection{Problem formulation}
The goal of trajectory prediction is to learn the posterior distribution, $P(\mathbf{Y}|\mathbf{X},\mathbf{I})$, of multiple vehicles’ future trajectories, $\mathbf{Y}=(Y_1, Y_2, ..., Y_N)$, given their past trajectories and properties (e.g. length and width), $\mathbf{X}=(X_1, X_2, ..., X_N)$, and lane information, $\mathbf{I}=(I_1, ..., I_M)$, where $N$ and $M$ mean the number of the vehicles and the number of lanes, respectively. The positions of all the vehicles are observed from time 1 to $t_{obs}$, and their future positions are predicted for time $T_{obs+1}$ to $T_{pred}$. The past trajectory and properties of the vehicle i can be written as $X_i = (x_{i,1}, ..., x_{i, T_{obs}}, p_{i,1}, ..., p_{i,k})$ where k is the number of properties considered. Also, the future trajectory of the vehicle $i$ can be written as $Y_i = (y_{i,T_{obs+1}}, …, y_{i, T_{pred}})$. The each element of the trajectory is a 2D position vector. The positions of the surrounding vehicles are represented in ego-vehicle based relative coordinate system. As the previous studies (e.g. \cite{xie2017vehicle, lee2017desire, li2019grip}), it is assumed that the position of the surrounding vehicles can be tracked from $t=1$ to $t=T_{obs}$.


\begin{figure}
    \centering
    \includegraphics[width=0.6\linewidth]{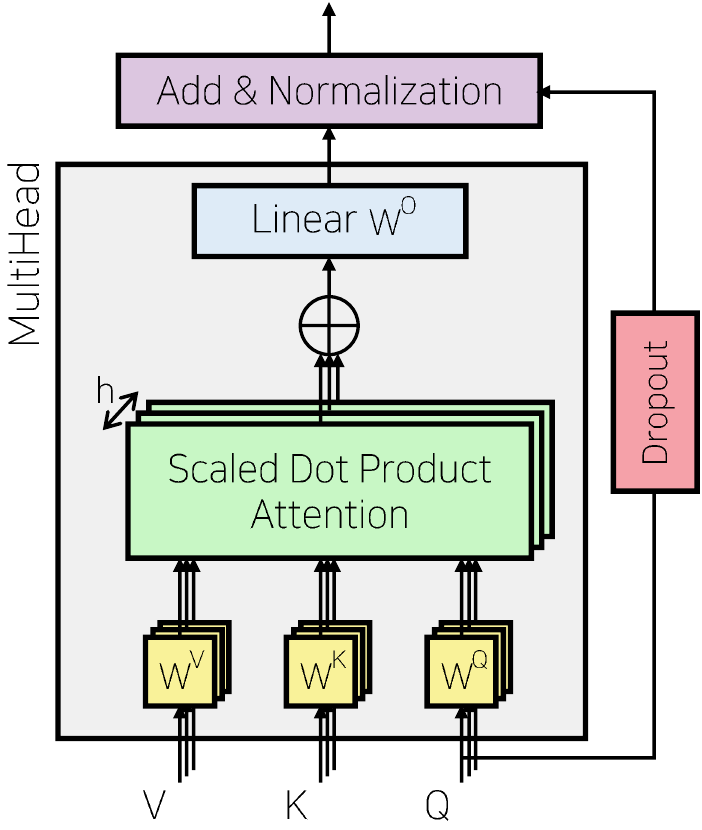}
    \caption{Structure of the attention layer for both the lane and the vehicles.}
    \label{fig:vehicle_attention}
\end{figure}

\subsection{Prediction model architecture}

The prediction model has an encoder-decoder structure. Both encoder and decoder use the multi-head attention, and the output of decoder is modeled as Gaussian distribution. The overall architecture is shown in Fig. \ref{fig:architecture}. Here, the encoder maps the past trajectories, $\mathbf{X}$, and lane information, $\mathbf{I}$, to a compressed representation, $\mathbf{Z}=(Z_1, ..., Z_N)$. Given $\mathbf{Z}$, the decoder generates predictive mean, $\mu$, and predictive covariance, $\Sigma$, for future trajectories, $\mathbf{Y}$. 

There are two attention layers in the proposed prediction model; vehicle attention layer and lane attention layer. Each attention layer have same architecture as shown in Fig. \ref{fig:vehicle_attention}. The attention layers map a set of queries, $Q$, a set of keys, $K$, and a set of values, $V$, into an output vector. The only difference between vehicle attention layer and lane attention layer is an input configuration. The vehicle attention layer uses vehicle embedding for $Q$, $K$ and $V$. However, the lane attention layer uses vehicle embedding as $Q$ and lane embedding as $K$ and $V$. Inside the attention layer in Fig. \ref{fig:vehicle_attention}, there is \textit{Scaled Dot Product Attention} layer. It enables the model to discover inter-dependencies within inputs. The attention computation in a single scaled dot product attention can be written as (\ref{eq:single_attention}). A set of queries, $Q$, is compared to a set of keys, $K$ by computing dot product attention, $QK^T$. The attention matrix can be obtained by scaling the dot product attention by $\frac{1}{\sqrt{d_k}}$ and normalizing it using softmax function. 

\begin{equation}
    \mathrm{Attention}(Q,K,V)=\underbrace{{\mathrm{softmax} \left( \frac{QK^T}{\sqrt{d_k}} \right)}}_{attention\, matrix} V
    \label{eq:single_attention}
\end{equation}

Multi-head attention performs the scaled dot product attention function in parallel for $h$ times. The independent attention outputs are concatenated and linearly transformed into the same dimension of $Q$. Each attention layer adopts a residual connection with dropout layer \cite{srivastava2014dropout} and a layer normalization \cite{ba2016layer}. 

\begin{equation}
\begin{split}
    \mathrm{MultiHead}(Q,K,V)&=\mathrm{Concat}(\mathrm{head_1}, ..., \mathrm{head_h})W^O \\
    \text{where}\quad\mathrm{head_i}&=\mathrm{Attention}(QW_i^Q, KW_i^K, VW_i^V)
\end{split}
\end{equation}

The encoder has both vehicle attention layer and lane attention layer. These attention layers generate attention based representations for each surrounding vehicle. The lane attention layer encodes the lane information related with the past trajectories of the surrounding vehicles. The use of the lane attention layer improves the prediction accuracy compared to simply the use of lane information as an embedding vector. The vehicle attention layer encodes the relations among the past trajectories of the vehicles. These attention outputs are concatenated as final encoder output, $\mathbf{Z}$. 

The decoder generates probabilistic prediction for future trajectories. The decoder consists of a single vehicle attention layer, where $Q$, $K$ and $V$ are encoder output $\mathbf{Z}$. It gathers the encoded information to predict the trajectories of the surrounding vehicles. The outputs of the decoder are predicted mean, $\mu=(\mu_1, ..., \mu_N)$, and predicted covariance, $\Sigma=(\Sigma_1, ..., \Sigma_N)$. 

\subsection{Implementation details}
The proposed prediction network is implemented in Python using Tensorflow \cite{abadi2016tensorflow}. The core parameters used for trajectory prediction are explained below. In the encoder, embedding of vehicles and lane information are performed first, and then, the embedded vectors are used in vehicle attention layer and lane attention layer, respectively. The embedding dimension for the past trajectories is 16 and the embedding dimension for the lane information is 4. In addition, the resulting output dimensions after the linear transformation using $W_i^Q, W_i^K, W_i^V$ are 8 and 32 for the vehicle attention layer and lane attention layer, respectively. Layer normalization with $\epsilon = 1\mathrm{e}{-6}$ is used for the output value of multi-head attention. The residual network, which adds the input and output of attention, used a dropout rate of 0.7 to prevent excessive use of residual connections during the training process. The vehicle attention layer used in the decoder is designed with the same parameters as those used in the encoder.

The following two loss terms are used in training attention based encoder-decoder: negative log likelihood loss in (\ref{eq:nll}) and reconstruction loss in (\ref{eq:recon_loss}). The total loss is weighted sum of two losses. 

\begin{equation}
    L_{NLL} = -\frac{1}{N} \sum_{i\in N} \log\left( \mathcal{N} (Y_i | \mu_i, \Sigma_i) \right)
    \label{eq:nll}
\end{equation}

\begin{equation}
    L_{Recon} = \frac{1}{N} \sum_{i\in N} \big\| Y_i - \mu_i \big\|
    \label{eq:recon_loss}
\end{equation}

For optimization, Adam optimizer \cite{kingma2014adam} is applied. The learning rate is set to 0.001. In this work, the batch size is set to 128.

\section{Experiments}
 In this section, experimental results are obtained based on publicly available highway dataset: highD dataset \cite{highDdataset} which is a large-scale naturalistic vehicle trajectory dataset from German highways observed by drones. NGSIM \cite{NGSIMdataset} is one of the largest datasets of naturalistic vehicle trajectories and widely used for trajectory prediction researches, but raw NGSIM trajectories should be carefully refined as the dataset contains erroneous trajectories such as false-positive collisions. To improve the quality of the dataset, the proposed algorithm used the highD dataset, instead of the NGSIM. The prediction model described in this work is running on a desktop PC equipped with a quadcore Intel Core i7-6700K CPU and an NVIDIA GeForce GTX 1080Ti GPU.

\newcommand{\ra}[1]{\renewcommand{\arraystretch}{#1}}

\begin{table}[]
\centering
\caption{Performance comparison of augmentation methods for test dataset.}
\ra{1.25}
\begin{tabular}{@{}ccccccc@{}}
\toprule
 & \multicolumn{3}{c}{\begin{tabular}[c]{@{}c@{}}Longitudinal \\ position error (m)\end{tabular}} & \multicolumn{3}{c}{\begin{tabular}[c]{@{}c@{}}Lateral \\ position error (m)\end{tabular}} \\ \cmidrule(l){2-7} 
Prediction horizon & 1s & 2s & 3s & 1s & 2s & 3s \\ \midrule
Linear & 0.71 & 1.67 & 3.41 & 0.17 & 0.55 & 1.31 \\
V-LSTM & 0.72 & 1.94 & 3.81 & 0.13 & 0.31 & 0.65 \\
ED-LSTM & 0.69 & 1.77 & 3.21 & 0.14 & 0.32 & 0.58 \\ \midrule
Proposed (h=2) & 0.59 & 0.77 & 1.31 & 0.08 & 0.14 & 0.30 \\
Proposed (h=4) & 0.43 & 0.47 & 0.89 & 0.04 & 0.06 & 0.11 \\
Proposed (h=8) & 0.54 & 0.58 & 1.09 & 0.07 & 0.11 & 0.18 \\ \bottomrule
\end{tabular}
\label{tab}
\end{table}

\begin{figure*}[t]
    \centering
    \begin{subfigure}{1\textwidth}
       \centering
       \includegraphics[width=17.4cm]{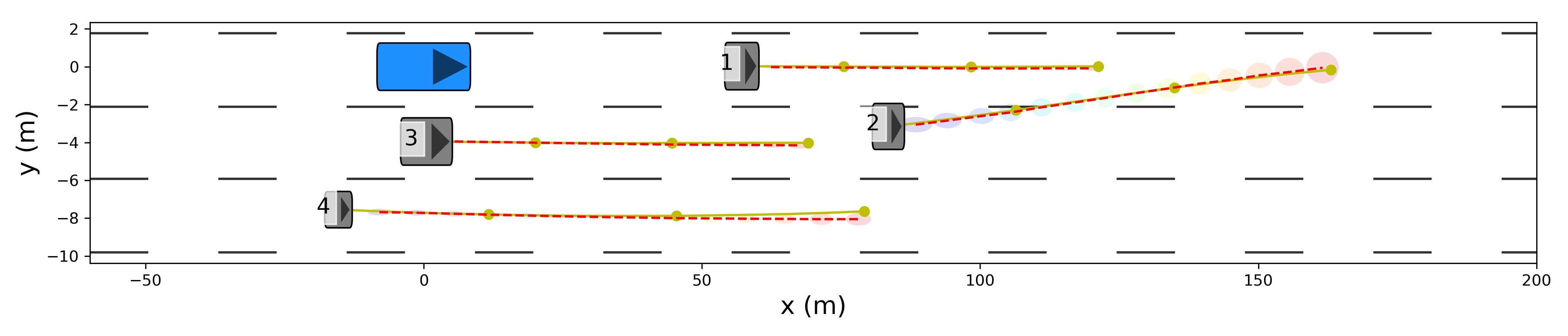}
    \end{subfigure}
    \caption{An example of trajectory prediction with four surrounding vehicles. The blue vehicle indicates ego-vehicle. Yellow solid line with three dots indicates true future trajectory, where dots represent positions at 1 second interval. Red dashed line indicates predicted future trajectory. Their uncertainties are drawn as ellipses from blue color to red color in chronological order. The boundaries of ellipses correspond to 3$\sigma$.}
    \label{fig:attn_trajectory} 
\end{figure*}

\subsection{Prediction model evaluation}

The proposed model provides a Gaussian distribution of future trajectories. To evaluate this, we used the root mean square error (RMSE) for the predicted mean value as the evaluation metric. In Table. \ref{tab}, we compare the performance of our model with some baseline methods:

\begin{itemize}
    \item \textit{Linear model (Linear)}: Extrapolating trajectory with assumption of linear velocity using an off-the-shelf Kalman filter.

    \item \textit{Vanilla LSTM (V-LSTM)}: Predicting future trajectory as a point estimates using an LSTM model. The past trajectory of a single vehicle is feeded to an LSTM.
    
    \item \textit{Encoder-decoder LSTM (ED-LSTM)}: LSTM encoder - LSTM decoder architecture is used for future trajectory prediction.
    
    \item \textit{Proposed model}: The proposed prediction model with various number of attention heads, $h$.
\end{itemize}

Even though the highway dataset is used for evaluation, the longer we predict, the larger error linear model provides. Especially, positional errors in the longitudinal direction is larger than the errors in the lateral direction. The vanilla LSTM performs better in terms of lateral position error than the linear model because it has the ability to predict the future trajectory based on the past trajectory. The ED-LSTM outperforms linear model and V-LSTM. 

However, the proposed algorithm shows much better performance than the baseline methods including ED-LSTM. There is a slightly different prediction performance depending on the number of heads, $h$, used in the proposed network. Generally, the model with 4 prediction heads tends to have smaller prediction errors than the model with 2 heads or 8 heads. The proposed model has an error value of 0.89m in the longitudinal direction and 0.11m in the lateral direction after 3 seconds. The proposed prediction model with high accuracy is expected to help autonomous vehicles drive safely.

\subsection{Analyzing the attention in trajectory prediction}
\begin{figure}[t]
    \centering
    \begin{subfigure}[b]{0.42\textwidth}
       \centering
       \includegraphics[width=6cm]{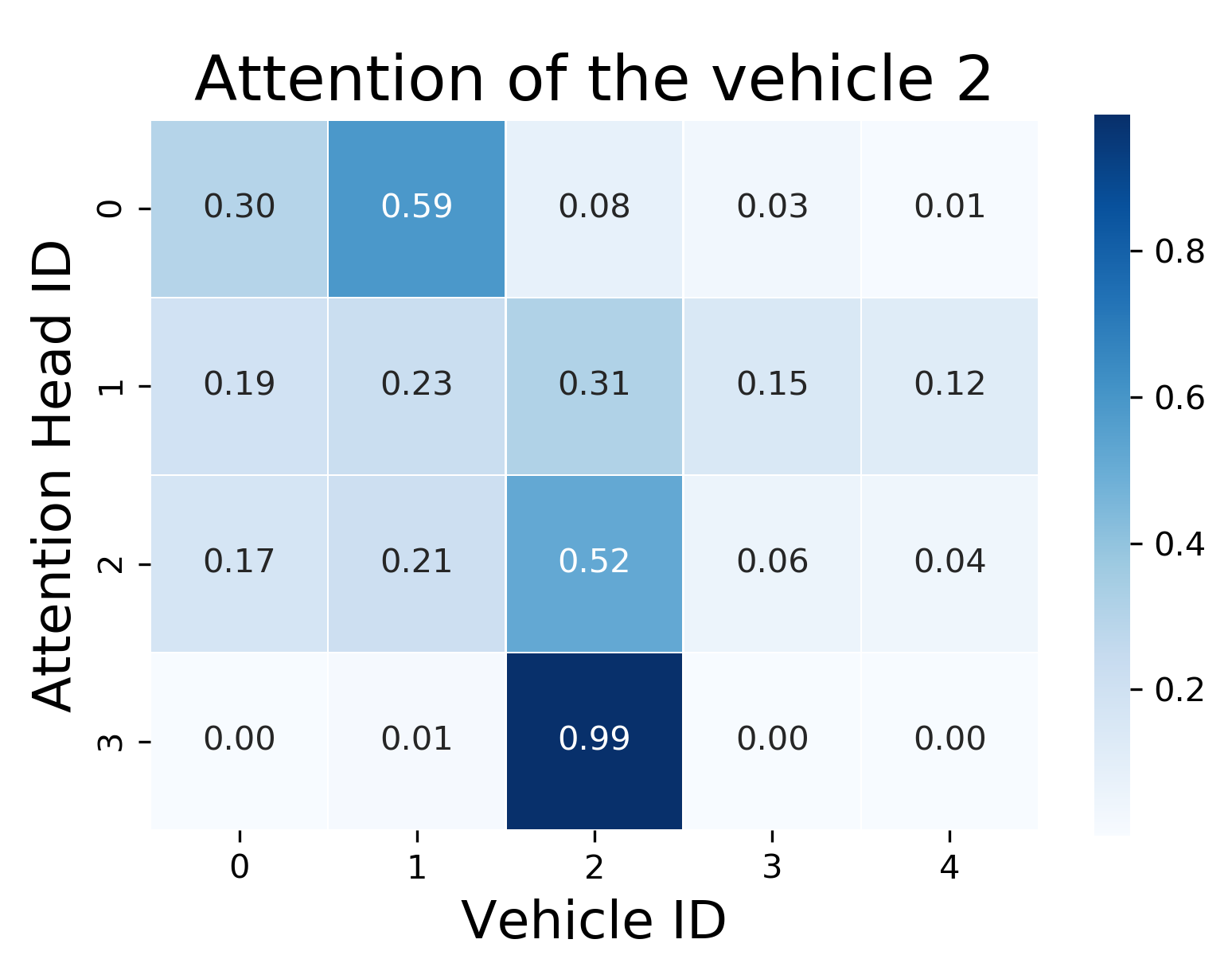}
       \label{fig:attn2}
       \caption{}
    \end{subfigure}
    
    \begin{subfigure}[b]{0.42\textwidth}
       \centering
       \includegraphics[width=6cm]{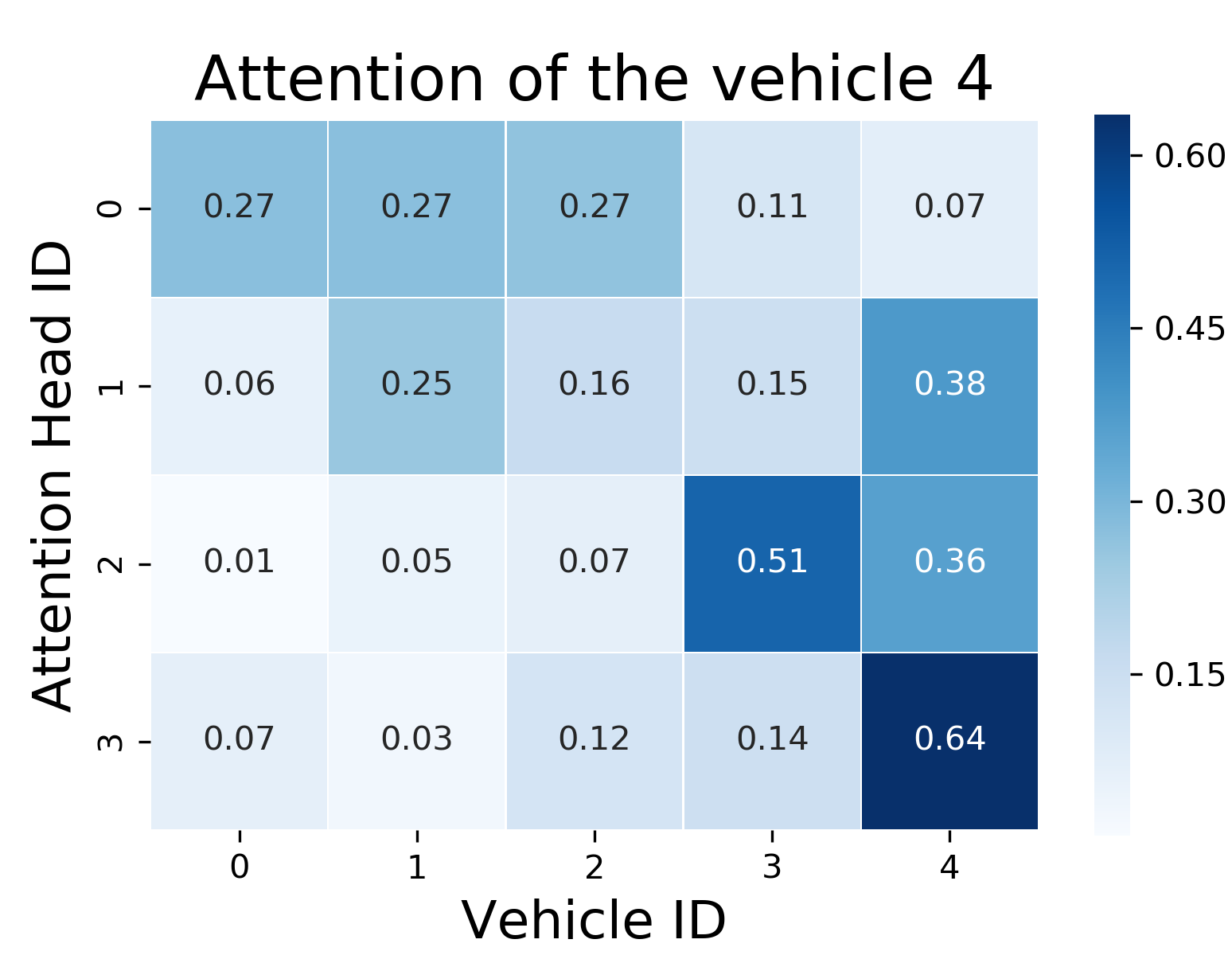}
       \label{fig:attn4}
       \caption{}
    \end{subfigure}

    \caption{The attention matrix for four attention heads in Fig. \ref{fig:attn_trajectory} situation. The ego vehicle ID is zero. (a) The vehicle 2 is predicted to change lane. (b) The vehicle 4 is predicted to keep the lane.}
    
    \label{fig:attention_example}

\end{figure}

One of the advantages of the proposed prediction algorithm is that it can learn attention during training. These Attention matrices can be used as an indicator of how strongly the interactions among vehicles occur. Fig. \ref{fig:attn_trajectory} is an example of predicting future trajectories of four surrounding vehicles. Note the Vehicle 2 changing lane and the Vehicle 4 keeping the lane. The attention matrix for these two vehicles is shown in Fig. \ref{fig:attention_example}. In the case of changing lanes, the prediction model attends to the vehicles (ego vehicle and the vehicle 1) in the lane to be changed as shown in Fig. \ref{fig:attention_example} (a). On the other hand, in the case of simply maintaining a lane, attention toward itself is higher than attention to other vehicles as shown in Fig. \ref{fig:attention_example} (b). In most cases, the last attention head has a high weight on itself, indicating that it depends heavily on its own past trajectory to predict future trajectory.

\subsection{Scalability for different number of the vehicles}
Unlike RNN variants, outputs of the encoder and decoder can be calculated in parallel by using Multi-head attention. This not only saves computation time in considering the interaction, but also has the advantage of being independent with the order of vehicles entering the network or the number of vehicles to be predicted, as long as the capacity of the memory allows.

\begin{figure*}
    \begin{subfigure}{1\textwidth}
      \centering
      \includegraphics[width=17.4cm]{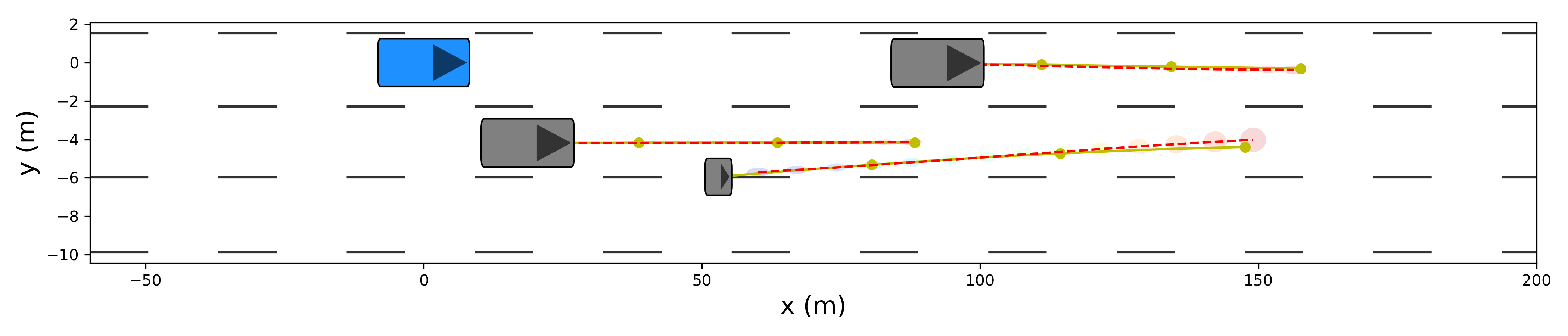}
    \end{subfigure}
    
    \begin{subfigure}{1\textwidth}
       \centering
       \includegraphics[width=17.4cm]{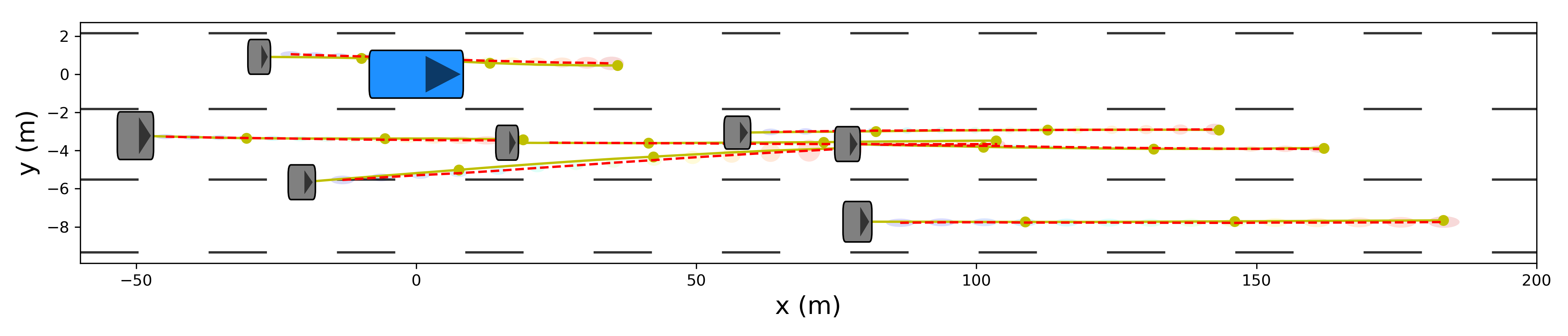}
    \end{subfigure}
    
    \begin{subfigure}{1\textwidth}
       \centering
       \includegraphics[width=17.4cm]{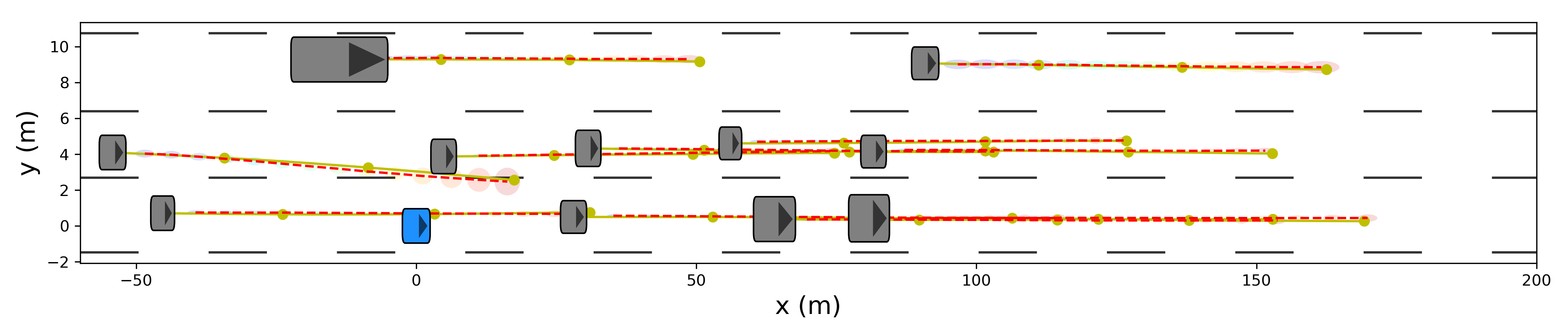}
    \end{subfigure}
    
    \begin{subfigure}{1\textwidth}
       \centering
       \includegraphics[width=17.4cm]{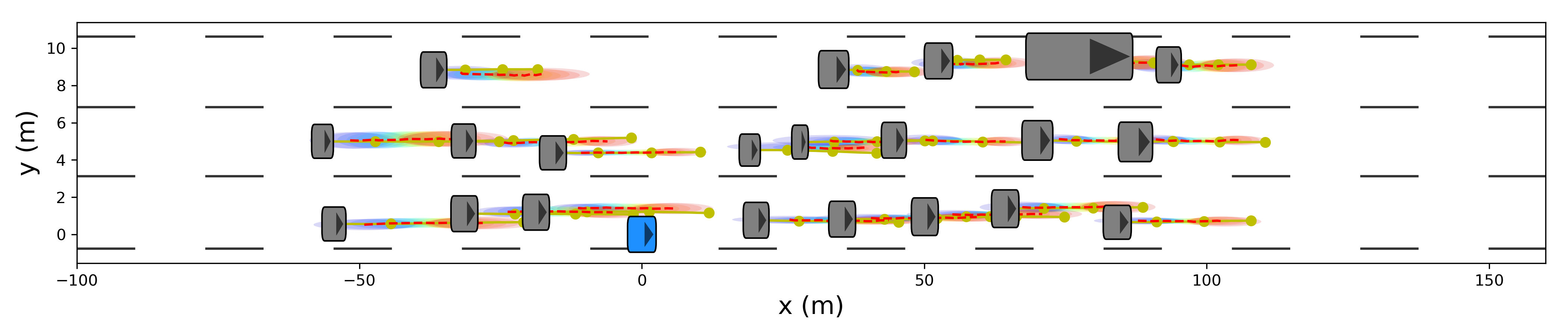}
    \end{subfigure}

    \caption{Prediction results for robustness test on scalability. The number of the surrounding vehicles is 3, 7, 11, 21 from top to bottom figures. The prediction network is trained only up to 10 surrounding vehicles. The blue vehicle indicates ego-vehicle. Yellow solid line with three dots indicates true future trajectory, where dots represent positions at 1 second interval. Red dashed line indicates predicted future trajectory. Their uncertainties are drawn as ellipses from blue color to red color in chronological order. The boundaries of ellipses correspond to 3$\sigma$.}
    \vspace{-120mm}
    
    \label{fig:some-prediction-results}

\end{figure*}

In order to validate its scalability, we first trained the prediction network in a scene with up to 10 surrounding vehicles. After that, the prediction performance of the network is tested for the future trajectories. The results are shown in Fig. \ref{fig:some-prediction-results}. The top two plots of Fig. \ref{fig:some-prediction-results} are the results of the prediction when there are fewer than 10 vehicles (3, 7 respectively), whereas the lower two are the results with more than 10 vehicles (11, 21 respectively). If there are 11 surrounding vehicles, the traffic density is not significantly different compared with 7 surrounding vehicles.  Because of this, even if the network has not learned the case of 11 surrounding vehicles, it can be confirmed that the prediction results are not very different from the actual future trajectory and has small uncertainty. In contrast, when there are 21 vehicles, the traffic density will be drastically changed from learned density. At this time, the mean of the predicted trajectory does not make a big difference from the actual trajectory, but its uncertainty tends to increase noticeably. This experiment demonstrates that the proposed network can predict the relatively robust trajectory even for different numbers of vehicles. In addition, the uncertainty generated when predicting 21 vehicles is epistemic uncertainty resulting from model uncertainty, and this uncertainty can be reduced by training the network with the trajectory of surrounding vehicles having similar velocity distribution.

\section{Conclusions}
In this paper, the multi-head attention based prediction model is proposed for future trajectory prediction of multiple vehicles considering the interactions. The proposed model has an encoder-decoder architecture, which incorporates the past trajectories and the lane information by vehicle attention layer and lane attention layer. The proposed methods is compared based on experimental data of the naturalistic trajectories at highway, and the evaluation results show that the proposed method outperforms the baseline methods. Additionally, the trained attention shows that the prediction network intuitively gathers information from a few influential vehicles to make better predictions. The learned distribution of vehicle trajectories can be used as constraints or costs for trajectory planning framework, which is our future research topic.

\section*{Acknowledgment}
This work was supported by the Technology Innovation Program (20006862, Development of Automated Driving Systems and Evaluation) funded By the Ministry of Trade, Industry \& Energy (MOTIE, Korea). The authors thank the anonymous reviewers for making this a better manuscript.

\addtolength{\textheight}{-12cm}   


\bibliographystyle{IEEEtran}
\bibliography{mybibfile}

\end{CJK}
\end{document}